\documentclass[a4paper]{article}

\usepackage[english]{babel}
\usepackage[utf8]{inputenc}
\usepackage{amsmath}
\usepackage{graphicx}
\usepackage[colorinlistoftodos]{todonotes}
\usepackage{amssymb,amsthm}
\usepackage{hyperref}
\usepackage{verbatim}
\usepackage{multirow}
\usepackage{caption}
\usepackage{hyperref}
\usepackage{authblk}
\usepackage{lipsum}

\newtheorem{theorem}{Theorem}
\newtheorem{definition}{Definition}

\newcommand\independent{\protect\mathpalette{\protect\independenT}{\perp}}
\def\independenT#1#2{\mathrel{\rlap{$#1#2$}\mkern2mu{#1#2}}}

\title{Reconciling Causality and Statistics \\
\begin{large}
An Introduction to the Do-Calculus with Examples
\end{large}
}

\author[1]{Pirmin Lemberger\thanks{Corresponding author: \texttt{p.lemberger@groupeonepoint.com}}}
\author[2]{Denis Oblin\thanks{Corresponding author: \texttt{denis.oblin@gmail.com}}}
\affil[1]{onepoint, 29 rue des Sablons, 75116 Paris}

\date{\today}

\begin{document}
\maketitle

\newcommand{\bx}{\mathbf{x}}
\newcommand{\by}{\mathbf{y}}
\newcommand{\bz}{\mathbf{z}}
\newcommand{\xfeatures}{x_1 \dots, x_p}
\newcommand{\Xset}{\mathcal{X}}
\newcommand{\Yset}{\mathcal{Y}}
\newcommand{\R}{\mathbb{R}}
\newcommand{\E}{\mathbb{E}}
\newcommand{\Do}{\mathrm{do}}

\begin{abstract}
Statisticians have warned us since the early days of their discipline that experimental correlation between two observations by no means implies the existence of a causal relation. The question about what clues exist in observational data that could informs us about the existence of such causal relations is nevertheless more that legitimate. It lies actually at the root of any scientific endeavor. For decades the only accepted method among statisticians to elucidate causal relationships was the so called Randomized Controlled Trial. Besides this notorious exception however, causality questions remained largely taboo for many. One reason for this state of affairs was the lack of an appropriate mathematical framework to formulate such questions in an unambiguous way. Fortunately thinks have changed these last years with the advent of the so called Causality Revolution initiated by Judea Pearl and coworkers. The aim of this pedagogical paper is to present their ideas and methods in a compact and self-contained fashion with concrete business examples as illustrations.
\end{abstract}

\section{A Stormy Relationship}
Consider the following three pairs of concomitant events: ``the sun rises shortly after the first crow of the rooster'', ``the number of visit to some website tends to increase when an ad banner is displayed in some appropriate location'' and ``the rate of skin cancer parallels the sale of ice creams on the French Riviera''. While common sense dictates that there is little chance to reduce cancer by limiting ice cream sales or to speed up sunrise by tickling a rooster so it crows earlier, it is not unlikely though that trying to optimize the position of an ad banner might indeed be worth the effort. So it seems our intuition about how the world works tells us something about this concomitant events which goes beyond what mere statistics can describe. Statistics indeed only deals with correlation, meaning it quantifies how likely it is that an event B will occur assuming that we have observed another event A. It tells us nothing however about what happens if we act on a system and it even provides no means to describe what an intervention on a system is. The conclusion is thus straightforward: in general we need more than just probability and statistics to express and answer causality questions. The method of \textbf{Randomized Controlled Trials} (RCT) which entirely rests on statistic analysis and is widely recognized as the gold standard for clinical trial is one noteworthy exception that we shall have more to say about later on.
\\

The innate inability of plain statistics to properly deal with causality in all its generality has only been recognized at the turn of this century \cite{WHY}. Before that, a controversy had been going on for nearly one century. We won't enter these lengthy debates because fortunately nowadays there is no reason for controversy anymore! The pioneering work of Judea Pearl \cite{WHY, CAU, EMP}, a philosopher and statistician, has put a definitive end to these quarrels by providing a mathematically sound framework that allows both to formulate causality questions in an unambiguous way and to answer them systematically whenever this is possible \cite{EMP}. Judea Pearl is largely responsible for a complete reshaping of the causality debate, which has since then been termed the \textbf{Causal Revolution}. For this and his other outstanding achievements on causal reasoning in AI he was awarded the Turing prize in 2011.
\\

\begin{figure}[h]
\centering
\includegraphics[scale=0.3]{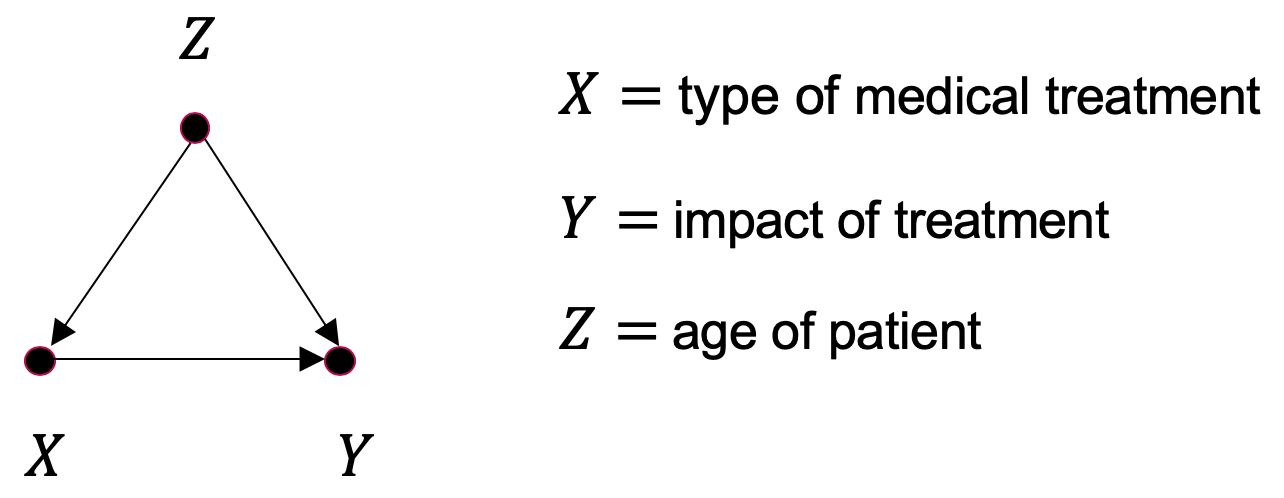}
\caption{\small Suppose $X$ designates the choice to administer a medicine to a patient, $Y$ is the state of health of that patient and $Z$ his or her age. If a doctor chooses to whom he or she administers the treatment, then the age $Z$ could become a confounder impacting both the choice $X$ of the treatment and the health state $Y$.}
\label{confounder_fig} 
\end{figure}

Before we dive into the subtleties of Judea Pearl's conceptions of causality, let's acknowledge that many data scientist who use \textbf{supervised machine learning} as their favorite tool never bother about causality issues. As a matter of fact they could could run into serious trouble because of this but it is nevertheless a legitimate attitude in two common situations that we briefly mention. 
\begin{itemize}
\item First, it could be that discovering a correlation between two type of events has value in its own right. For instance learning that users who  bought product A are also likely to buy product B can provide some insight into a population of customers. The important point is that any subsequent action driven by such a correlation based prediction should not retro-act in any way on the measured variables.

\item Second, there are other situations where a passive observation of some feature $X$ has indeed the same effect (on predicting a target variable $Y$) as an intervention where we prescribe the value of that feature. Imagine for instance that $X$ is the rotation speed of an air pump and $Y$ is the pressure it maintains in a room. Once we have learned the relation between $X$ and $Y$ by passive observations, we will be able to predict the air pressure when we set the speed of the pump. Intuitively, the causal relationship from $X$ to $Y$ is direct and unperturbed. We will precisely characterize theses situations in section \ref{DoCalc_section} when we describe the do Calculus.

\end{itemize}
Where Judea Pearl's approach really shines is in less obvious situations where we would like to infer the effect of setting $X=x$ on the value of $Y$ in the \linebreak presence of a whole network of causal relationships that involves other variables $Z$. These other variables are often called \textbf{concomitants} \cite{COX} and may be observed or not. In its simplest version the root of the difficulty in answering causal questions is depicted in figure \ref{confounder_fig} which illustrates the notion of a \textbf{confounder} $Z$.
In intuitive terms a confounder $Z$ is a variable that could potentially hinder the identification of a causal relationship between two other variables $X$ and $Y$ by being a cause of both. An observed change in the value of $Y$ that results from a change of $X$ could just as well be attributed to a change of $Z$. Yet, if had access to the values of the confounder $Z$, we could perhaps adjust for its effect and thus extract the desired effect of $X$ on $Y$. The procedure for doing this, when it is possible, is well known to statisticians and goes under the name ``\textbf{controlling} for $Z$''. In a nutshell, what Pearl's machinery achieves is a vast generalization of this controlling procedure:
\\

\definecolor{antiquewhite}{rgb}{0.98, 0.92, 0.84}
\noindent\fcolorbox{gray}{antiquewhite}{
\minipage[c]{\dimexpr1.0\linewidth-2\fboxsep-2\fboxrule\relax}
Pearl's graphic and algebraic tools allow the identification of situations in which we may possibly disambiguate a causal relationship between two variables $X$ and $Y$. Wherever this is possible these tools provide explicit formulas for computing the effect on a variable $Y$ of an intervention that prescribes the value $x$ of another variable $X$. These formulas merge two kinds of information. First, they uses the values of some concomitants $Z$ which are just observed. Second, they leverage a causal graph $G$ which encapsulates our knowledge of the causal relationships between the variables within a system of interest. 
\endminipage}
\\

Later we shall consider still another way to interpret Pearl's work namely as a vast generalization of RCT's.
\\

Pearl's conceptual and computational tools have several important merits:
\begin{itemize}

\item From a practical point of view, they allow to \textbf{anticipate the effect of an intervention on a system without actually having to perform any action on it}. This is obviously extremely useful in situations where we cannot afford to act on a system, whether for ethical or financial reasons.

\item Still from a practical point of view, they all to \textbf{integrate causal prior knowledge} about a system with measurements in a fully consistent manner.

\item From conceptual stand point, they provide clear definitions of what is meant by a causal influence and by an intervention, as we shall see shortly. This \textbf{settles the secular causality controversy} and reconciles statistics with causality.

\end{itemize}
The sequel of this article is organized as follows. Section 2 introduces the main concepts and tools defined by Pearl and coworkers. After a brief review of \textbf{Bayesian networks} which allows us to introduce the notion of \textbf{$d$-separation}, we define what a \textbf{functional causal model} is and describe the extent to which it can be inferred from data. We define the notions of \textbf{intervention} and of \textbf{identifiabiliy}. We explain the graphical \textbf{back-door} and \textbf{front-door} criteria that allow to identify causal effects in some specific circumstances. At last we introduce the rules of the \textbf{do-calculus} that are even more flexible. Section 3 discusses various business examples as illustrations. The last section offers some concluding remarks and tries to explain why in our opinion these tools still remain little known within the data scientist community at large.

\section{The Causal Revolution in a Nutshell}
\label{CR_in_Nutshell_section}
In this section we will present the main \textbf{definitions} and \textbf{mathematical results} underpinning the Causal Revolution. We define what \textbf{causal relationships} are, how and when we can identify them from data. Then we define what an \textbf{intervention} is and describe the graphical and algebraic tools which allow to identify situations where we can compute the consequences of an intervention using data collected from pure observation. For proofs we refer to Pearl's book on Causality \cite{CAU} and to the original papers \cite{EMP}.

\subsection{Bayesian Networks Recap}
The tools we shall present in the following subsections rely heavily on a graphical method called the \textbf{$d$-separation criterion} that allows answering general independence questions among groups of random variables (r.v). These tools were pioneered by Pearl who developed the concept of a \textbf{Bayesian Network}. There are many good textbooks on this topic \cite{BIS} but we provide a brief introduction both to introduce notations and to make the presentation reasonably self-contained. Readers familiar with this material are invited to jump to subsection \ref{Causal_Model_subsection}.
\\

We use upper case letters like $X, Y$ or $Z$ to designate r.v. and boldface upper case to denote sequences (or sets) of r.v $\mathbf{X}=(X_1, X_2,..., X_m)$, with lower indices to denote the components of $X$. For simplicity we shall assume that each component takes only a finite number of values. The corresponding realizations of these r.v. will be denoted with corresponding lower case letters $\mathbf{x}=(x_1, x_2,..., x_m)$.
\\

\begin{figure}[h]
\centering
\includegraphics[scale=0.28]{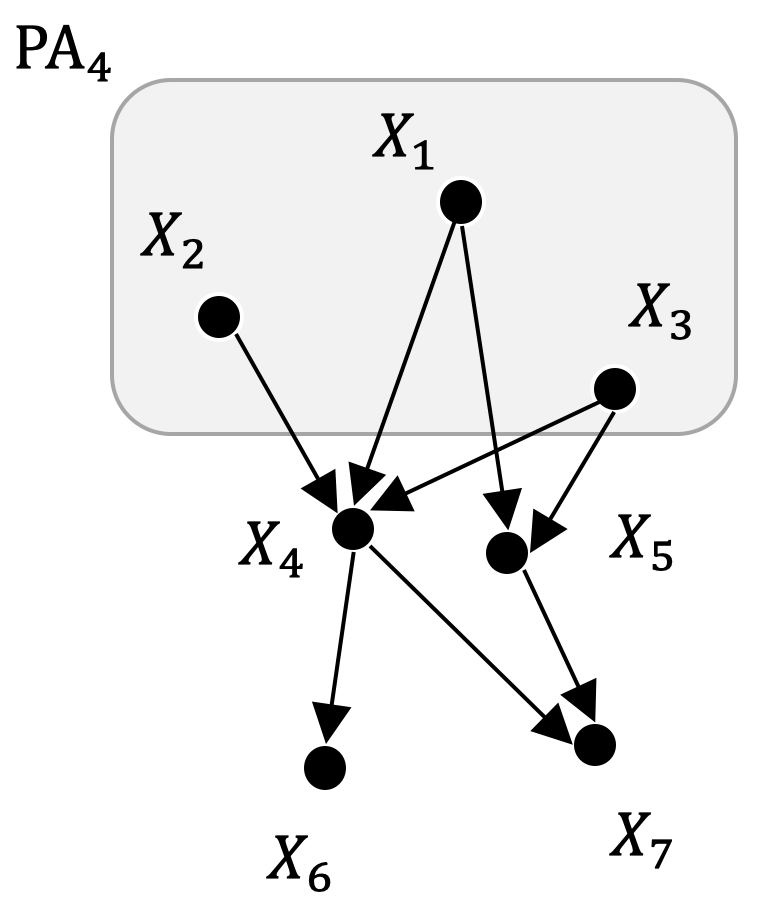}
\caption{\small A unique DAG $G$ can be associated to a probability distribution $p$ on an ordered set or r.v. $X_1,...,X_n$. $\mathrm{PA}_4$ are the parent variables of $X_4$}
\label{DAG_fig} 
\end{figure}

The first step is to associate a graph $G$ to a probability distribution $p(\mathbf{x})=p(x_1, x_2,..., x_m)$. Iterating the product rule for conditional probabilities we have:
\begin{eqnarray}
	\label{prob_fact}
	p(\bx)
	       =    p(x_1, x_2,...,x_n) 
	      &=& p(x_1)\:p(x_2,...,x_n|x_1) \nonumber\\	
		  &=& p(x_1)\:p(x_2|x_1)\:p(x_3,...,x_n|x_1, x_2) \nonumber\\
		  &=& ... \nonumber\\
		  &=& \prod_{j=1}^n p(x_j|x_1,...,x_{j-1})
\end{eqnarray}
where the conditioning set for $j=1$ is empty. Now if $X_j$ is only conditioned on a subset of $X_1,...,X_{j-1}$, let's denote the minimal such subset by $\mathrm{PA}_j$, the \textbf{parents} of $X_j$, so that we can write:
\begin{equation}
	\label{prod_decomp}
	p(x_1, x_2,...,x_n) = \prod_{j=1}^n p(x_j|\mathrm{pa}_j).
\end{equation}
A natural representation of (\ref{prod_decomp}) is to use a \textbf{Bayesian network} which is a \textbf{directed acyclic graph} (DAG) where each node of the graph corresponds to a variable $X_j$ and a directed link points from each member of $\mathrm{PA}_j$ towards $X_j$ as figure \ref{DAG_fig} shows. 
We say that the probability distribution $p$ and the graph $G$ are \textbf{compatible}. Note also that according to (\ref{prob_fact}) the DAG associated to a generic distribution $p$ with no specific factorization properties is a fully connected graph. Hence all information in a DAG is really contained it the missing links. 
\\

For a given order of the r.v. this graph $G$ is unique. However, if we had chosen another order for the r.v. in $p$ we would have obtained another graph $G'$. Therefore, the natural question that arises is to determine when two DAG $G$ and $G'$ \textbf{observationally equivalent}. More precisely we want to ask whether any probability distribution $p$ which is compatible with $G$ is also compatible with another graph $G'$. The following can be shown \cite{CAU}:
\begin{theorem}[observationnal equivalence]
\label{obs_equiv_theorem}
Two DAG's are observationally equivalent if two conditions are met. First the graphs that result from stripping of the arrows on the edges of $G$ and $G'$ (their skeletons) should be the same. Second, $G$ and $G'$ should have the same $\nu$-structures, which are nodes with converging arrows whose tails are not connected by an arrow.
\end{theorem} 
Let $\mathbf{X}, \mathbf{Y}$ and $\mathbf{Z}$ be three groups of r.v. whose joint distribution is $p$. Of particular interest is the question whether $\mathbf{X}$ is independent of $\mathbf{Y}$ conditionally on observing $\mathbf{Z}$, that is whether 
\begin{equation}
	\label{indep_1}
	p(\mathbf{x},\mathbf{y}|\:\mathbf{z})=
	p(\mathbf{x}|\mathbf{z}) \:
	p(\mathbf{y}|\mathbf{z}),
\end{equation}
or, equivalently, whether
\begin{equation}
	\label{indep_2}
	p(\mathbf{x}|\:\mathbf{y}, \mathbf{z}) = p(\mathbf{x}|\mathbf{z}).
\end{equation}
When the answer is yes, we denote this by $\mathbf{X} \independent \mathbf{Y} \:|\: \mathbf{Z}$. The $d$-separation theorem below allows us to answer this question graphically by looking at the graph $G$ associated to $p$. For this we need the following graphical concept:
\begin{definition}[$d$-separation]
A path $\gamma$ (a sequence of consecutive edges oriented in any direction) in a DAG $G$ is said to be blocked by a set of conditioning (or observed) variables $\mathbf{Z}$ if and only if one of the following two conditions is met:
\begin{enumerate}
\item $\gamma$ contains a \textbf{chain} $i \rightarrow m \rightarrow j$ or a \textbf{fork} $i \leftarrow m \rightarrow j$ such that $m$ belongs to $Z$.
\item $\gamma$ contains a \textbf{collider} $i \rightarrow m \leftarrow j$ such that neither $m$ nor any of its descendants (as defined by the arrows on the edges in $G$) belongs to $\mathbf{Z}$.
\end{enumerate}
A set $\mathbf{Z}$ is said to d-separate $\mathbf{X}$ from $\mathbf{Y}$ whenever $\mathbf{Z}$ blocks every path from a node in $\mathbf{X}$ to a node in $\mathbf{Y}$. 
\end{definition}
\begin{figure}[h]
\centering
\includegraphics[scale=0.28]{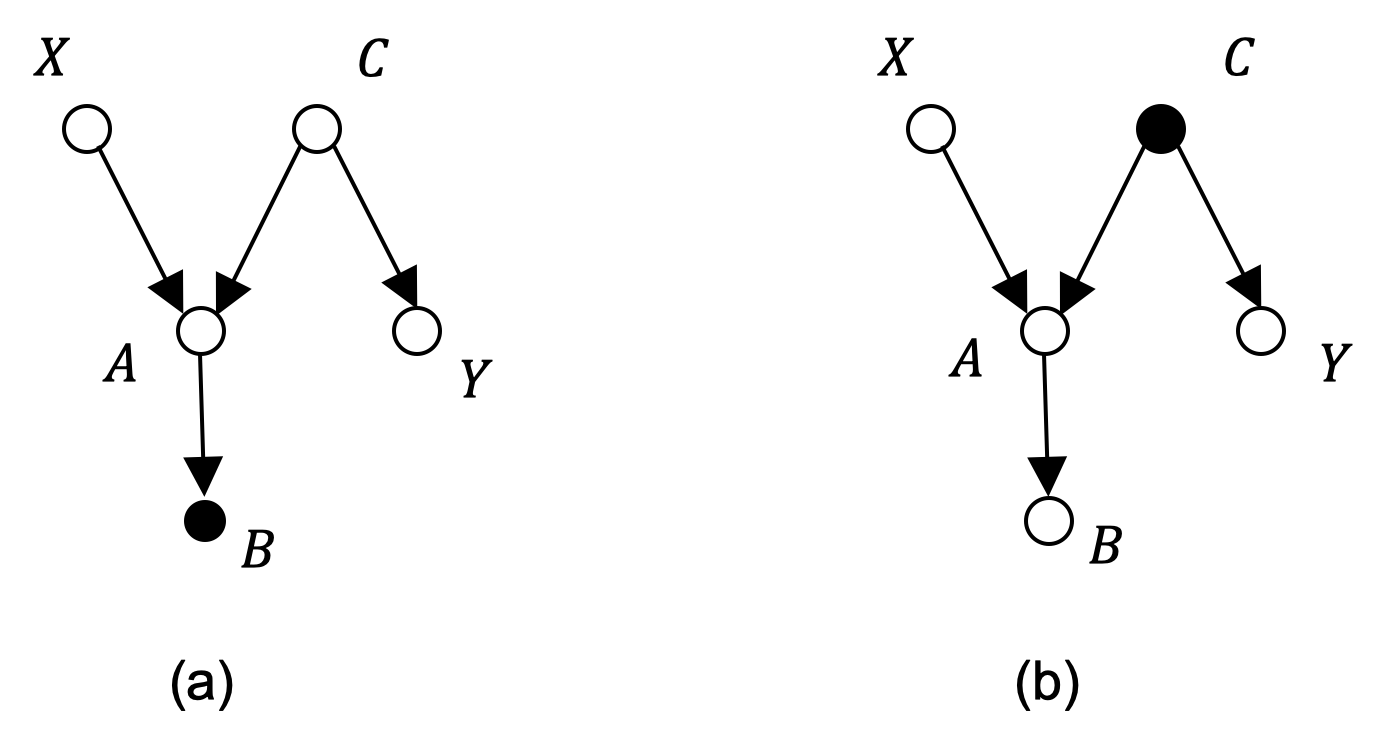}
\caption{\small Conditioning variables are depicted as filled circles. In (a) the path from $X$ to $Y$ is neither blocked by $B$, because it is an observed descendant of the collider $A$, nor is it by $C$ which is an unconditioned fork. In (b) the path from $X$ to $Y$ is blocked by $C$ because it is an conditioned fork.}
\end{figure}
We can now formulate the $d$-separation criterion which relate the factorization properties of $p$ described by a compatible DAG $G$  and the independence relationships they imply:
\begin{theorem}[$d$-separation theorem]
\label{d-sep_theorem}
Let $\mathbf{X},\mathbf{Y},\mathbf{Z}$ be three disjoint sets of nodes in a DAG $G$ compatible with a distribution $p$. Then the corresponding r.v. distributed according to $p$ satisfy $\mathbf{X} \independent \mathbf{Y}\:|\:\mathbf{Z}$ whenever $\mathbf{X}$ and $\mathbf{Y}$ are $d$-separated by $\mathbf{Z}$.
\end{theorem}
The converse is true as well. An important consequence of theorem \ref{d-sep_theorem} is that the set of conditional independence among r.v. is determined by the \textbf{topology} of $G$ only, while the ordering of the r.v. in $G$ plays no role \cite{CAU}. When we consider the $d$-separation criterion as a graphical test on the nodes of a DAG $G$ we shall write $\left( \mathbf{X} \independent \mathbf{Y}\:|\:\mathbf{Z}\right)_G$. Using (\ref{indep_2}) we can then rewrite theorem \ref{d-sep_theorem} compactly as
\begin{equation}
	\label{d-separation_rewrite}
	p(\mathbf{x}|\:\mathbf{y}, \mathbf{z}) = p(\mathbf{x}|\mathbf{z})
	\textnormal{  when  }
	\left( \mathbf{X} \independent \mathbf{Y}\:|\:\mathbf{Z}\right)_G \textnormal{  for  } G \textnormal{  compatible with } p.
\end{equation}
We insist that that so far there has been no discussion of causality, only independence questions were examined. There is however an easy \textbf{mnemonic} for the $d$-separation criterion if we momentarily allow ourselves to associate a causal meaning to the arrows in $G$. Indeed, if we fix the value of the middle node $m$ in a chain or a fork we block the information flow between the two end nodes $i$ and $j$ which thus become independent. If, on the contrary, we observe the value of a collider node $m$, or of some of its descendants, we actually observe a common consequence of $i$ and $j$. Knowing both $i$ and $m$ for example will thus tell us something about $j$, which thus opens the information flow. This independence analysis based on $d$-separation prefigures the genuine causal analysis we develop in the next subsections.

\subsection{Functional Causal Models and Interventions}
\label{Causal_Model_subsection}
\subsubsection*{Structural Equations}
\begin{figure}[h]
\centering
\includegraphics[scale=0.15]{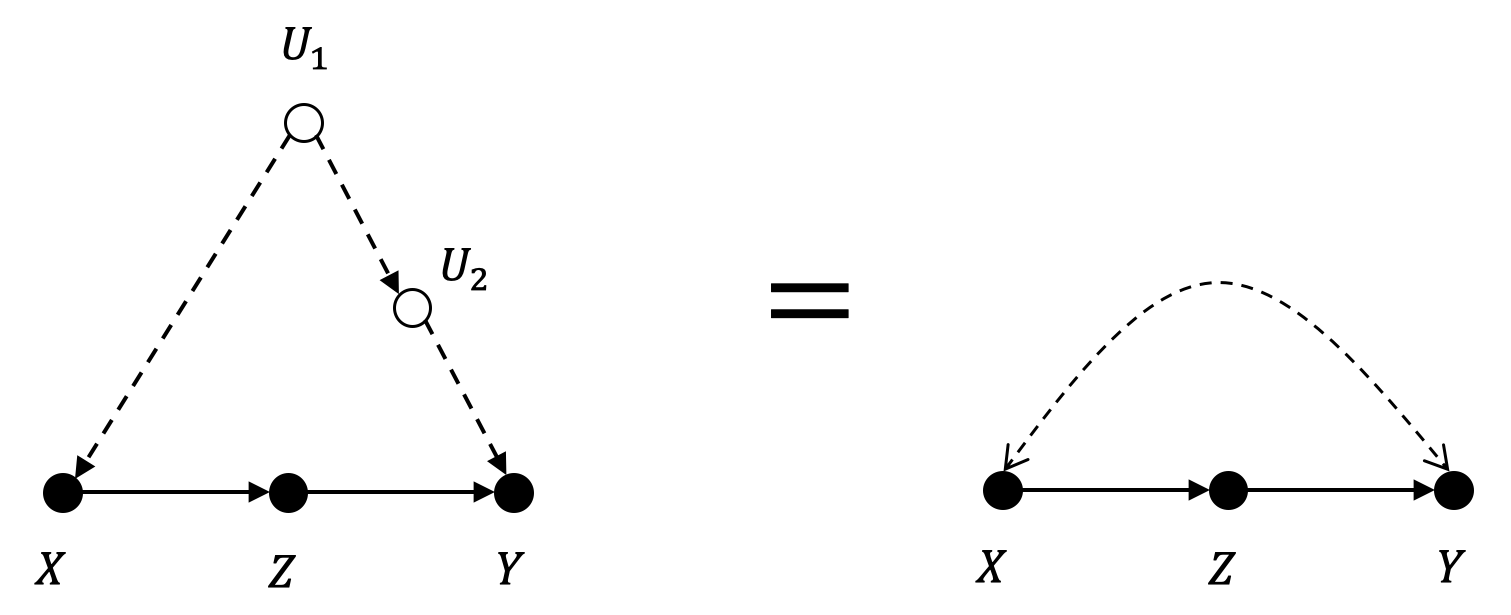}
\caption{\small A causal graph $G$ where observed r.v. like $X,Y$ or $Z$ are represented as full circles, while unobserved r.v. like $U_1$ or $U_2$ are represented as empty circles. A confounding arc is a path which joins two observed r.v., contains only unobserved r.v. and has no converging arrows.}
\label{causal_ex_1_fig}
\end{figure}

A conditional probability $p(\mathbf{y}|\mathbf{x})$ tells us how the probability of seeing $\mathbf{Y}=\mathbf{y}$ is affected by having observed that $\mathbf{X}=\mathbf{x}$. It tells us nothing however about what happens when we prescribe the value of $\mathbf{X}$. To answer that question we need more information than $p$ contains, namely information about how the probabilistic model $p$ will change under an external intervention. This is precisely what a \textbf{functional causal model} $M$ does. It is defined as a set of functional relationships which describe how each r.v. $X_j$ is determined as a function of the others and of some noisy disturbances. It thus explains how the data is generated. Using the notation $\mathrm{PA}_j$ to denote the set of r.v. that directly influence $X_j$, a causal model is then defined as a set of \textbf{structural equations}: 
\begin{equation}
\label{struct_eq}
	X_j=f_j(\mathrm{PA}_j, \epsilon_j),\;\; j=1,...,n,
\end{equation}
where the $f_j$'s are deterministic functions of the parents variables $\mathrm{PA}_j$ and of some noisy disturbance $\epsilon_j$. An important point is that the $\epsilon_j$ should be independent. In other words, each relation is assumed to be disturbed by a single random perturbation. If a perturbation were to affect several r.v. simultaneously, we should promote it to an unobserved $X_j$ with a structural equation describing its influence on other variables. A \textbf{causal graph} $G$ can be associated to a causal model $M$ as shown in figure \ref{causal_ex_1_fig} which also defines the notion of a \textbf{confounding arc}.
\\

The probability distribution $p$ on a set of r.v. $\mathbf{X}$ is completely specified by the model $M$ which comprises the set of deterministic functions $f_j$ and the distributions of the disturbances $\epsilon_j$. In practice these will however remain unspecified. Beware that the links in figure \ref{DAG_fig} represent conditional dependencies $p(x_j|\mathrm{pa}_j)$ while they represent deterministic dependencies $x_j=f_j(\mathrm{pa}_j, \epsilon_j)$ in figure \ref{causal_ex_1_fig}.
\\

Assume that we are given a causal graph $G$ associated to a causal model $M$ which generates a probability distribution $p$. What independence relationships hold for this $p$ ? The answer is as simple as we could have dreamed: they are precisely those determined by the $d$-separation criterion applied to $G$. This follows from the fact that $p$ is compatible with $G$ in the Bayesian network sense and from the $d$-separation theorem\footnote{Following the proof of theorem 1.4.1 in \cite{CAU}, note that (\ref{struct_eq}) implies that the joint distribution $p(x_1,...,x_n,\epsilon_1,...,\epsilon_n)$ on the original r.v. and on the disturbances is compatible, in the Bayesian network sense, with the graph $G'$ constructed from $G$ by adding to it all nodes corresponding to the disturbances $\epsilon_j$ together with their links directed towards the nodes $X_j$. The same paths $\gamma$ in $G$ which are blocked in $G$ will also be blocked in $G'$. The statement thus directly follows from the $d$-separation theorem \ref{d-sep_theorem} applied to $G'$.}.

\subsubsection*{The do-Operator}
\begin{figure}[h]
\centering
\includegraphics[scale=0.4]{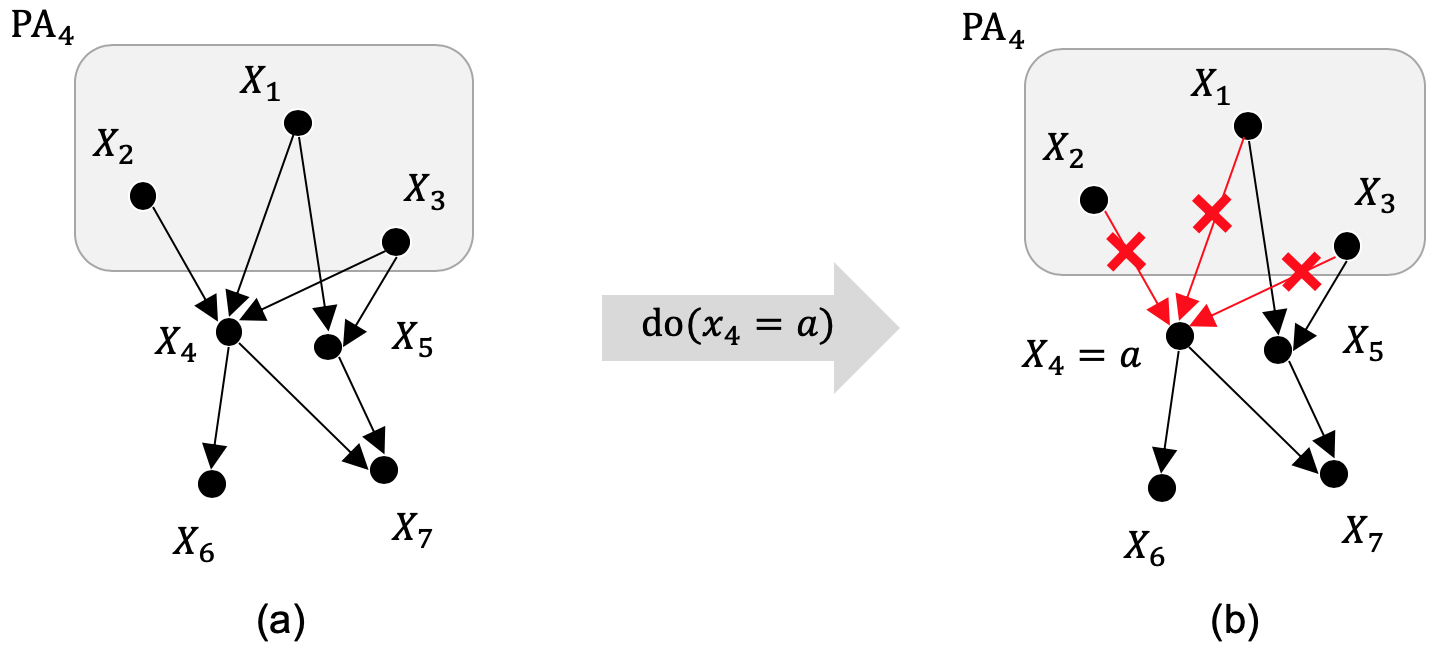}
\caption{\small Removing the links between the variable $X_4$ and its parents $\mathrm{PA}_4$ to define an intervention. If (a) represents $p(\bx)$, then (b) represents $p(\bx|\Do(x_4=a))$.}
\label{set_value_graph}
\end{figure}

An \textbf{intervention} on a system described by a causal model is an alteration of the structural equations (\ref{struct_eq}) where some relationships are replaced by others. In the simplest case where we fix the value of say $X_i$ to $a$ this amounts to replacing the equation $X_i=f_i(\mathrm{PA}_j, \epsilon_i)$ with $X_i=a$. The corresponding causal graph of the modified model is thus obtained from $G$ by removing all links connecting the nodes to $X_i$ to their parents $\mathrm{PA}_i$ as figure \ref{set_value_graph} shows.
\\
After the intervention, the probability distribution $p(\mathbf{x})$ is replaced by a new one which we denote by $p(\mathbf{x}|\:\Do(x_i=a))$. This defines the \textbf{do-operator}. Using this definition and the factorization (\ref{prob_fact}) for the distribution $p$ we can also express the post intervention distribution as a truncated factorization:
\begin{equation}
	\label{do_op_def_eq}
	p(\bx|\:\Do(x_i=a)):=\left\{
		\begin{array}{ll}
			\displaystyle{\frac{p(\bx)}{p(x_i|\:\mathrm{pa}_i)} = \prod_{j\neq i}}\:p(x_j|\:\mathrm{pa}_j)  &\text{ if } x_i = a,\\
			\\        		
        		0 &\text{ if } x_i\neq a.
    		\end{array}
	\right.
\end{equation}
In general $p(\bx|\:\Do(x_i=a))$ is of course different from the classical conditional $p(\bx|\:x_i=a)$. One trivial case where both expressions are indeed identical is when $X_i$ has no parents ($\mathrm{PA}_i=\varnothing$). The general question of when an action on a variable $X$ has the same consequence on another variable $Y$ as an observation will be answered precisely when we present the do-calculus in section \ref{DoCalc_section}.
\\

To get more intuition about the do-operator let's compute the effect of an intervention $\Do(x_i=a)$ on a single variable $Y$ which is different from both $X_i$ and its parents $\mathrm{PA}_i$. Plugging 
\begin{equation}
	\label{cond_prob_eq}
	\frac{p(\bx)}{p(x_i|\:\mathrm{pa}_i)} =
	\frac{p(\bx)}{p(x_i,\mathrm{pa}_i)/p(\mathrm{pa}_i)} = 
	p(\bx|x_i,\mathrm{pa}_i)\:p(\mathrm{pa}_i)
\end{equation}
into (\ref{do_op_def_eq}) for $x_i=a$ and marginalizing over all variables except on $y$ and on $x_i$ we obtain
\begin{eqnarray}
	\label{cond_prob_eq2}
	p(y|\:\Do(x_i=a)) 
		& =& \sum_{\bx\setminus (y \cup x_i)} p(\bx|\:\Do(x_i=a))  \nonumber\\
		& = &\sum_{\bx\setminus (y \cup x_i) } p(\bx|x_i=a,\mathrm{pa}_i)\:p(\mathrm{pa}_i) \nonumber\\
		& = &\sum_{\mathrm{pa}_i} p(\mathrm{pa}_i)
		\sum_{\bx\setminus (y \cup x_i \cup \mathrm{pa}_i) } p(\bx|x_i=a,\mathrm{pa}_i) \nonumber\\
		& = &\sum_{\mathrm{pa}_i} p(\mathrm{pa}_i) \: p(y|x_i=a,\mathrm{pa}_i)
\end{eqnarray}
This conditioning and weighting operation on the variables $\mathrm{PA}_i$ to compute $p(y|\:\Do(x_i=a)) $ in (\ref{cond_prob_eq2}) is known as \textbf{adjusting} for the \textbf{direct causes} of $X_i$. 
\\


Equation (\ref{cond_prob_eq2}) shows that when the parents $\mathrm{PA}_i$ of a manipulated variable $X_i$ are measurable, then the effect on $Y$ of an intervention on $X_i$ can be computed from passive observations only. Indeed these determine both the prior $p(\mathrm{pa}_i)$ and the conditional $p(y|x_i=a,\mathrm{pa}_i)$. In practice we could for instance use machine learning for this purpose.
\\

The interesting question then is whether we can compute the effect of an intervention when this is not the case, that is when some of the r.v. in $\mathrm{PA}_i$ are \textit{not} measurable? More generally, the question that naturally arises is then the following: given a causal graph $G$ and a subset $\mathbf{Z}$ of observable concomitants can we express $p(y|\:\Do(x=a))$ in terms of the distribution on the observed variables $\mathbf{Z}, X$ and $Y$ only? When this is the case we shall say that the causal effect of $X$ on $Y$ is \textbf{identifiable}. Lets compactly summarize this concept of identifiability \cite{TEST_ID}. 
\\

\definecolor{antiquewhite}{rgb}{0.98, 0.92, 0.84}
\noindent\fcolorbox{gray}{antiquewhite}{
\minipage[c]{\dimexpr1.0\linewidth-2\fboxsep-2\fboxrule\relax}
Passive observation of the variables $X$ and $Y$ and a group $\mathbf{Z}$ observable concomitants informs us on their joint distribution $p(x,y,\mathbf{z})$, as long as we have enough data of course. However, the effect $p(y|\Do(x))$ on $Y$ of an intervention on $X$ is only encoded in the causal model $M$ which generates this $p$, but \emph{not} in $p$ itself! Thus the problem we face is that two different causal models $M_1$ and $M_2$ could generate the same $p$. To predict the effect of an intervention we thus need more information than is contained in $p$. The causal graph $G$ associated to the causal model $M$ can sometimes provide this missing information, \emph{without} the need for us to have a complete knowledge of the causal model $M$. When this is the case, we say that the intervention $p(y|\Do(x))$ is identifiable from $G$.
\endminipage}
\\

In subsections \ref{answering_with_graphs_section} and \ref{DoCalc_section} we shall describe graphical and symbolic tools to answer these questions. 

\subsubsection*{Interpreting Randomized Control Trials with the do-Operator}
Lets quickly make contact between the definition (\ref{do_op_def_eq}) of the do operator and the \textbf{Randomized Control Trial} (RCT) procedure. Consider a situation where we would like to assess the effectiveness of a treatment on a group of patients. Say that $X=1$ means we apply the treatment and $X=0$ we don't and that $Y$ represents the resulting impact of their health (``improved'', ``stable'', ``deteriorated''). Patients are characterized by features like $Z_1=$age, $Z_2=$gender and $Z_3=$``is diabetic'' which could also influence their health state. 
\begin{figure}[h]
\centering
\includegraphics[scale=0.24]{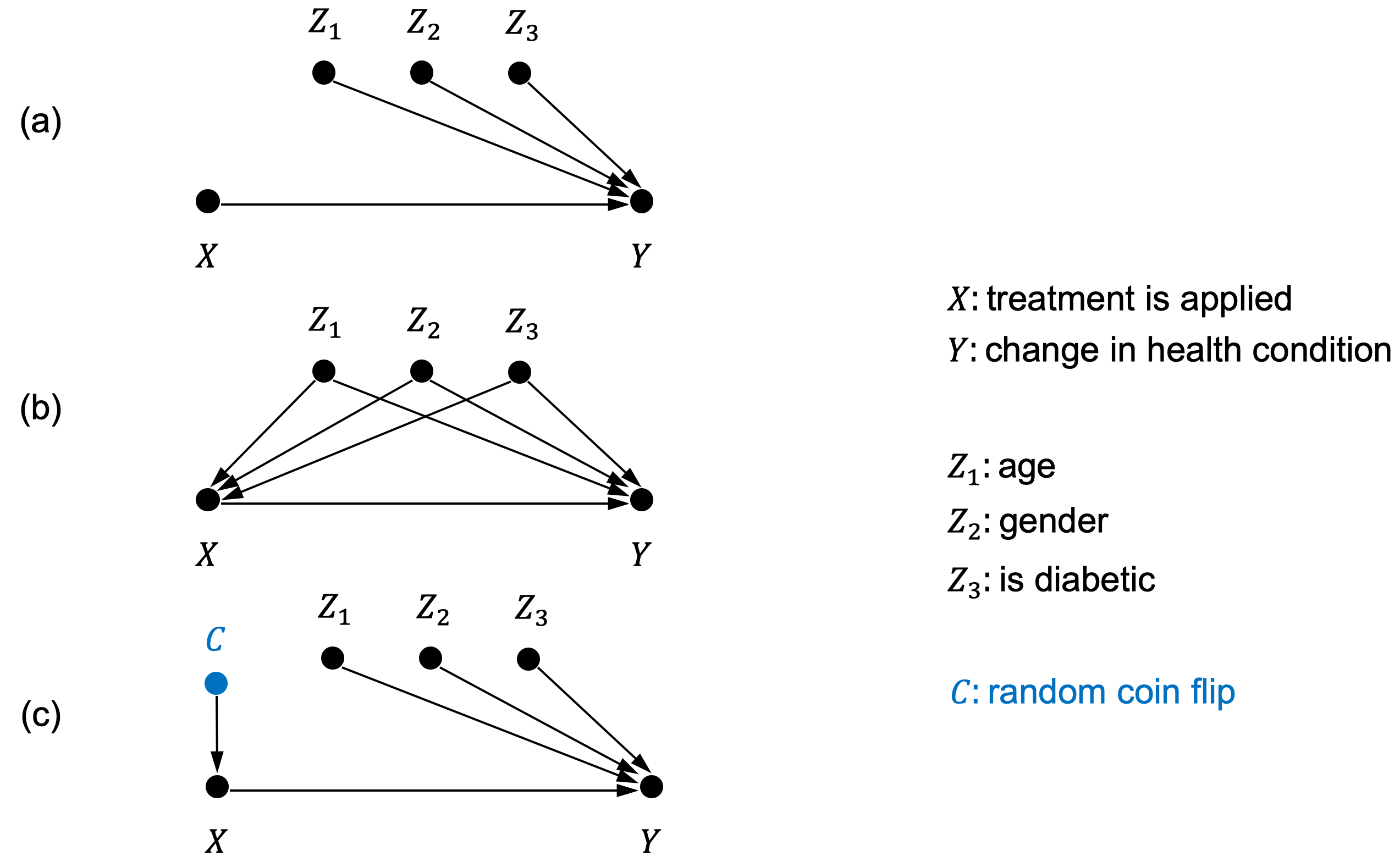}
\caption{\small{(a) The causal diagram of the combined impact of the treatment $X$ and the features $\mathbf{Z}=(Z_1=\text{age},Z_2=\text{gender},Z_3=\text{``is diabetic''})$ of a patient on his health $Y$, (b) the causal diagram when the experimenter chooses himself the patients to whom he administers the treatment, thus potentially transforming $\mathbf{Z}$ into confounders, (c) the treatment is administered by randomly flipping a coin $C$ thus decoupling the treatment from any potential confounders.}}
\label{RCT_fig}
\end{figure}
\\
The corresponding causal diagram is shown in figure \ref{RCT_fig} (a). For this diagram we would like to compute $p(y|\:\Do(x))$. If the experimenter chooses which patient she will administer the treatment to, she could inadvertently transform the features $\mathbf{Z}$ into confounders that will simultaneously impact the decision $X$ to administer the treatment and the health $Y$ of the patient. This is represented in figure \ref{RCT_fig} (b). The RCT procedure subordinates the decision to administer the treatment to  the result of flipping a perfect coin $C$ whose outcome is neither influenced by $\mathbf{Z}$ nor has any direct influence on $Y$. It thus cuts the same incoming links to the instrumented variable $X$ as those which had to be removed to define $p(y|\:\Do(x))$ in (\ref{do_op_def_eq}). In such a circumstance $p(y|x,c)=p(y|x)$, because $X$ blocks the path between $C$ and $Y$. Therefore (\ref{cond_prob_eq2}) implies that $p(y|\:\Do(x))=p(y|x)$. In other words, when using the RCT procedure observing $X$ has the same effect on predicting $Y$ as prescribing the value of $X$.

\subsubsection*{Can we Infer a Causal Graph from Raw Data?}
In sections \ref{answering_with_graphs_section} and \ref{DoCalc_section} we shall describe tools to answer the identifiability question above assuming that we are given a causal graph $G$ and that some of its variables $\mathbf{Z}$ can be measured. But what if we have no causal graph $G$ available in the first place? Can we infer it directly from data? This is secular question which has sparked a lot of controversy. In every day life we all use some clues to identify causal relationship, the most important one being the assumption that a cause always precedes its effects. Analyzing these temporal aspects of causality would however quickly drive us into discussing the origin of the asymmetry between the past and the future and other foundational issues in physics \cite{TIME}. As interesting as they are, we will not address them here, because this would take us too far away. We restrict the subject to analyzing the patterns of data which can identity functional causal relations defined by structural equations (\ref{struct_eq}). We refer to \cite{WHY} for a broader panorama of causality. 
\\

The common wisdom among statisticians has long been that this identifying a causal model from data is plain impossible. And, indeed, in general it is impossible as we shall confirm. The interesting point though is that contrary to this common belief there are practically relevant situations where this identification of $G$ \emph{is} possible, at least in part. Again, a thorough discussion of these matters would be out of the scope for this short paper. We therefore limit ourselves to summarizing one significant result in \cite{CAU} which shows that, assuming a specific interpretation of Occam's razor, an equivalence class of causal graphs $G$ can be inferred when all variables $\mathbf{X}$ in $G$ are measurable. 
\\

So lets assume that the variables $\mathbf{X}$ are all measurable and that their probability distribution is $p(\bx)$. With no further assumptions there could exist many different causal graphs $G$ compatible with $p$. Therefore we need to put  restrictions on the causal models $M$ which we assume have generated $p$. We shall indeed put two such restrictions. The first, \textbf{minimality}, conforms with standard scientific practice and restricts our search to the simplest possible causal graphs $G$ only, the second, \textbf{stability}, will be a condition on the causal model $M$ itself. To formulate them precisely we assume that the causal model $M$ defined by (\ref{struct_eq}), depends on \textbf{parameters} $\theta$ that define both the deterministic dependencies $f_j$ and the distributions of the noise variables $\epsilon_j$. Such a parameterized causal model $M$ therefore really defines a set of distributions $p$ that could be obtained by varying the parameters $\theta$ of $M$.
\begin{enumerate}

\item Given two causal graphs $G_1$ and $G_2$ consider the parameterized models $M_1$ and $M_2$ that are associated with these graphs. Whenever the set of distributions generated by $M_2$ contains those generated by $M_1$ we say that the causal graph $G_1$ is \textbf{preferable} to $G_2$. In other words, the set of distributions compatible with $G_1$ is smaller than the set compatible with $G_2$. A causal graph $G$ is said to be \textbf{minimal} among a set of graphs when no other graph in that set is preferable to $G$.

\item Given a distribution $p$ generated by a causal model $M$ with fixed parameters values $\theta$ we should make sure that $p$ has no accidental independence relationships that would hold only for those specific set of parameters values. Changing the value of $\theta$ to another set of values $\theta'$ should not destroy any independence relationship $X \independent Y|Z$ which hold for $p$. When this holds we say that the causal model generates a \textbf{stable} distribution \cite{CAU}.

\end{enumerate}
Theorem \ref{obs_equiv_theorem} on observational equivalence then implies the following \cite{CAU}:
\begin{theorem}
If a causal model $M$ generates a stable distribution $p$ then, up to observational equivalence, there is a unique minimal causal graph $G$ compatible with $p$.
\end{theorem}
A class of observationally equivalent graphs can be represent graphically by a representation called a \textbf{pattern} \cite{CAU}. A pattern is a partially directed DAG where some edges are directed while others are not. The directed ones correspond to edges that are common to all members of the pattern, while the undirected one could be oriented in either direction. Pearl and Verma \cite{IC_ALGO, PEVE} developed an algorithm, the \textbf{IC algorithm} (for Inferred Causality) which constructs the pattern associated to any stable distribution $p$ on a set of observable variables. We refer to \cite{CAU} for a detailed description of the IC algorithm and for various extensions of IC to causal models with latent variables. Other interesting results obtained by Pearl pertain to the falsifiability of some assumptions in a causal graph \cite{FAL1, FAL2}.

\subsection{Answering Causality Questions with Graphs}
\label{answering_with_graphs_section}
As we have seen earlier, the causal effect $p(y|\Do(x)$ is identifiable as soon as the parents $\mathrm{PA}(X)$ of the cause $X$ are measurable. Formula (\ref{cond_prob_eq2}) answers the question by showing how we should adjust the conditionals $p(y|x,\mathrm{pa})$ according to the prior probabilities $p(\mathrm{pa})$ of the direct causes. In this subsection we describe two simple graphical criteria which answer the question in more general cases.
\subsubsection*{The Back-Door Criterion}
The following holds \cite{CAU,EMP}:
\begin{theorem}
\label{backdoor_theorem}
The effect $p(y|\Do(x))$ on $Y$ of an intervention on $X$ is identifiable from the causal graph $G$ when a group $\mathbf{Z}$ of observable r.v. satisfy the following two conditions:
\begin{enumerate}
\item No variable $Z_i$ is a descendant of $X$ in $G$. In other words $X$ should have no influence on any of the variables in $\mathbf{Z}$.
\item The group of variables $\mathbf{Z}$ should block any path $\gamma$ between $X$ and $Y$ in $G$ that has an incoming arrow into $X$ (back-door paths).
\end{enumerate}
When both conditions are verified we have the reduction:
\begin{equation}
	\label{backdoor_reduction}
	p(y|\Do(x)) = \sum_\mathbf{z} p(y|x,\mathbf{z})\:p(\mathbf{z})
\end{equation}
\end{theorem}
The above reduction formula can be considered as a generalization of the basic adjustment formula (\ref{cond_prob_eq2}). Both conditions in theorem \ref{backdoor_theorem} can be easily tested systematically. Various sets $\mathbf{Z}$ can be tried to optimize the cost of their measurement.
\begin{figure}[h]
\centering
\includegraphics[scale=0.28]{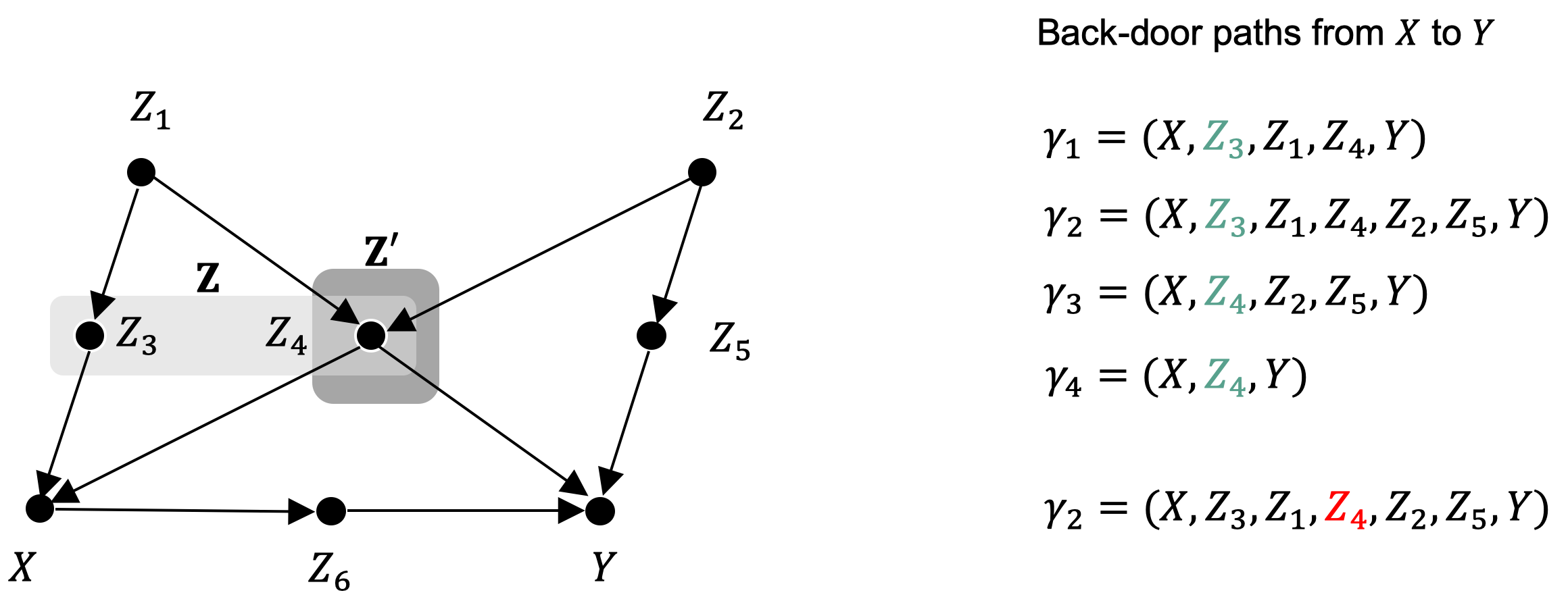}
\caption{\small{An example of a group of r.v. $\mathbf{Z}=(Z_3,Z_4)$ that make the intervention of $X$ on $Y$ identifiable. Indeed $Z_3$ is a measured chain that blocks $\gamma_1$ and $\gamma_2$ while $Z_4$ blocks $\gamma_3$ as a measured chain and $\gamma_4$ as a measured fork. On the other hand $\mathbf{Z}'=(Z_4)$ does not because the path $\gamma_2$ remains unblocked by $Z_4$ which is a measured collider.}}
\label{backdoor_fig}
\end{figure}
\\
Figure \ref{backdoor_fig} gives an example of a group $\mathbf{Z}=(Z_3,Z_4)$ which makes the causal influence of $X$ on $Y$ identifiable while $\mathbf{Z}'=(Z_4)$ does not.

\subsubsection*{The Front-Door Criterion}
Condition 1 in theorem \ref{backdoor_theorem} requires that the cause $X$ should not affect the observable variables $\mathbf{Z}$. Fortunately, another criterion, namely the Front-Door criterion, allows us to prove that the effect of $X$ on $Y$ is nevertheless identifiable in other cases as well. Consider the situation depicted in figure \ref{front_door_fig} obtained by amalgamating $Z_1$ to $Z_5$ in figure \ref{backdoor_fig} into a group $\mathbf{U}$ which we assume to be unobserved then renaming $Z_6$ as $Z$ and denoting $\mathbf{Z}=(Z)$. This variable $Z$ is under direct influence of $X$ and does not block the path $X\leftarrow \mathbf{U} \rightarrow Y$. Therefore $Z$ meet neither of the two back-door criteria. Note that the following three conditions hold in the example of figure \ref{front_door_fig}:
\begin{figure}[h]
\centering
\includegraphics[scale=0.28]{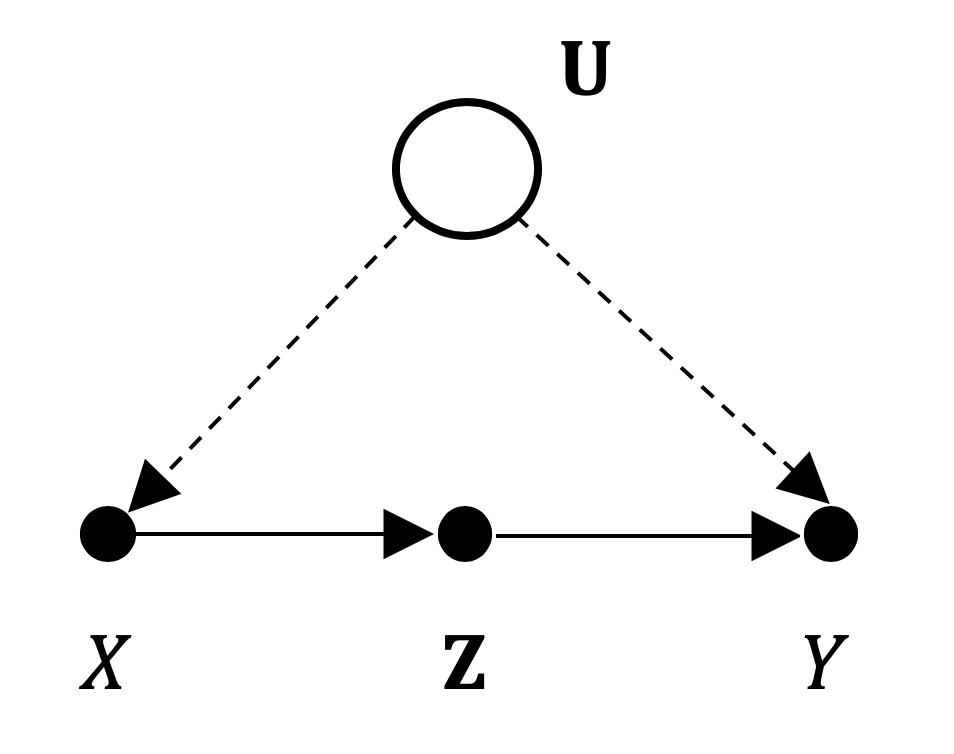}
\caption{\small{A causal graph which satisfies the conditions for applying the front-door criterion.}}
\label{front_door_fig}
\end{figure}
\begin{enumerate}
\item The variables $\mathbf{Z}$ block the directed path from $X$ to $Y$.
\item All back-door paths from $X$ to $\mathbf{Z}$ are blocked. Indeed in the example the path $X\leftarrow \mathbf{U} \rightarrow Y \leftarrow \mathbf{Z}$ is blocked at $Y$ which is an unconditioned collider.
\item All back-door paths from $\mathbf{Z}$ to $Y$ are blocked by $X$. Indeed in the example the path $\mathbf{Z}\leftarrow X \leftarrow \mathbf{U} \rightarrow Y$ is blocked by $X$ which is a conditioned chain.
\end{enumerate}
When these three conditions are met we say that the variables $\mathbf{Z}$ satisfie the \textbf{frond-door criterion} relative to the causal effect of $X$ on $Y$. The following then holds:
\begin{theorem}
\label{frontdoor_theorem}
When the variables $\mathbf{Z}$ satisfy the front-door criterion for the effect of a variable $X$ on a variable $Y$, then this effect is identifiable and is given by
\begin{equation}
	\label{front_door_eq}
	p(y|\:\Do(x)) = \sum_{\mathbf{z}} p(\mathbf{z}|x)\:\sum_{x'}p(y|x',\mathbf{z})\:p(x').
\end{equation}
\end{theorem}
Notice that the variables $\mathbf{u}$ do not appear in the above reduction formula as should be the case for unmeasured variables. This second graphical tool thus complements our arsenal for discovering identifiable causal graphs. We will prove it in subsection \ref{DoCalc_section} as an example application of the do calculus.

\subsection{Answering Causality Questions with the Do-Calculus}
\label{DoCalc_section}
Graphical tools are nice because they allow us to identify causal relationships without having to do any math. Unfortunately, this is not always possible. In this subsection we present three simple algebraic rules \cite{EMP} that can be combined to transform expressions containing both ordinary conditioning $.|\mathbf{x}$ and interventions $.|\Do(\mathbf{x})$ into equivalent expressions. These are the rules of the \textbf{do Calculus}. Our goal of course is to find an appropriate sequence of these transformation to eventually get rid of all $.|\Do(\mathbf{x})$ in order to prove that an intervention is identifiable. 
\\

Assume we are given four disjoint subsets of variables $\mathbf{X},\mathbf{Y},\mathbf{Z}$ and $\mathbf{W}$ in a causal graph $G$. Let us denote by $G_{\overline{\mathbf{X}}}$ and $G_{\underline{\mathbf{X}}}$ the graphs obtained from $G$ by respectively deleting the links incoming into $\mathbf{X}$ and those outgoing from $\mathbf{X}$. To represent simultaneous removal of incoming and outgoing links we use a notation like $G_{\overline{\mathbf{X}}\underline{\mathbf{Z}}}$. At last, $\mathbf{Z}(\mathbf{W})$ will denote the set of $Z$-nodes which are not ancestors of any $W$-nodes in $G_{\overline{\mathbf{X}}}$. 
\\

In the 3 rules below, $\mathbf{X}$ always enters as a $\Do(\mathbf{X})$ intervention, while $\mathbf{W}$ occurs as a passive observation and $\mathbf{Y}$ as the consequences we would like to predict. We also define a mixed conditional probability on $\mathbf{Y}$ given an intervention on $\mathbf{X}$ and an observation of $\mathbf{W}$:
\begin{equation}
\label{mixed_cond_eq}
	p(\mathbf{y}|\:\Do(\mathbf{x}), \mathbf{w}):=
	\frac{p(\mathbf{y},\mathbf{w}|\:\Do(\mathbf{x}))}{p(\mathbf{w}|\:\Do(\mathbf{x}))}.
\end{equation}

\begin{theorem}[do Calculus rules] Assume that the distribution $p$ is generated by a causal model associated with the graph $G$. Then the following holds:
\begin{itemize}
\item\textbf{Rule 1} [removing an observation]
	\subitem 
	$p(\mathbf{y}|\Do(\mathbf{x}), \mathbf{z}, \mathbf{w}))=p(\mathbf{y}|
	\Do(\mathbf{x}), \mathbf{w}))
	\textnormal{ when } 
	\left(\mathbf{Y} \independent \mathbf{Z}|\:\mathbf{X},
	\mathbf{W}\right)_{G_{\overline{\mathbf{X}}}}$.
\item \textbf{Rule 2} [removing an intervention]
	\subitem $p(\mathbf{y}|\:\Do(\mathbf{x}), \Do(\mathbf{z}), \mathbf{w}))=p(\mathbf{y}|\:\Do(\mathbf{x}), \mathbf{w}))$
	\textnormal{ when } $\left(\mathbf{Y} \independent \mathbf{Z}|\:\mathbf{X},
	\mathbf{W}\right)_{G_{\overline{\mathbf{X}},\overline{\mathbf{Z}(\mathbf{W})}}}$.
\item \textbf{Rule 3} [replacing an intervention with an observation]
	\subitem $p(\mathbf{y}|\Do(\mathbf{x}), \Do(\mathbf{z}), \mathbf{w}))=p(\mathbf{y}|\Do(\mathbf{x}), 		
	\mathbf{z},  \mathbf{w}))$
	\textnormal{ when } $\left(\mathbf{Y} \independent \mathbf{Z}|\:\mathbf{X},
	\mathbf{W}\right)_{G_{\overline{\mathbf{X}}\underline{\mathbf{ Z}}}}$.
\end{itemize}
\end{theorem}
These rules generalize the basic $d$-separation rule (\ref{d-separation_rewrite}) which we recall here for comparison:
\begin{equation}
	\label{classic-d-sep}
	p(\mathbf{y}|\:\mathbf{z}, \mathbf{w}) = p(\mathbf{y}|\mathbf{w})
	\textnormal{   when  }
	\left( \mathbf{Y} \independent \mathbf{Z}\:|\:\mathbf{W}\right)_G.
\end{equation}
In fact (\ref{classic-d-sep}) is a special case of rule 1 when $\mathbf{X}=\varnothing$. 
In the other way around, we can easily understand rule 1 as a consequence of (\ref{classic-d-sep}) when we remember that an intervention $\Do(\mathbf{x})$ simply removes from $G$ all links entering into $\mathbf{X}$. What results is precisely the graph $G_{\overline{\mathbf{X}}}$ used in rule 1. Rules 2 and 3 use sub-graphs of $G_{\overline{\mathbf{X}}}$. Rule 3 answers the important question of whether an intervention on the variable $Z$ has the same effect as its passive observation. The simplest case where rule 3 applies is when $G$ is the two-node graph $\mathbf{Z} \rightarrow \mathbf{Y}$, describing a direct causal effect, for which $G_{\underline{\mathbf{Z}}}$ reduces to two disconnected nodes thus trivially ensuring $(\mathbf{Y} \independent \mathbf{Z})_{G_{\underline{\mathbf{Z}}}}$.

\subsubsection*{An Example - Deriving the Front-Door formula}
\begin{figure}[h]
\centering
\includegraphics[scale=0.25]{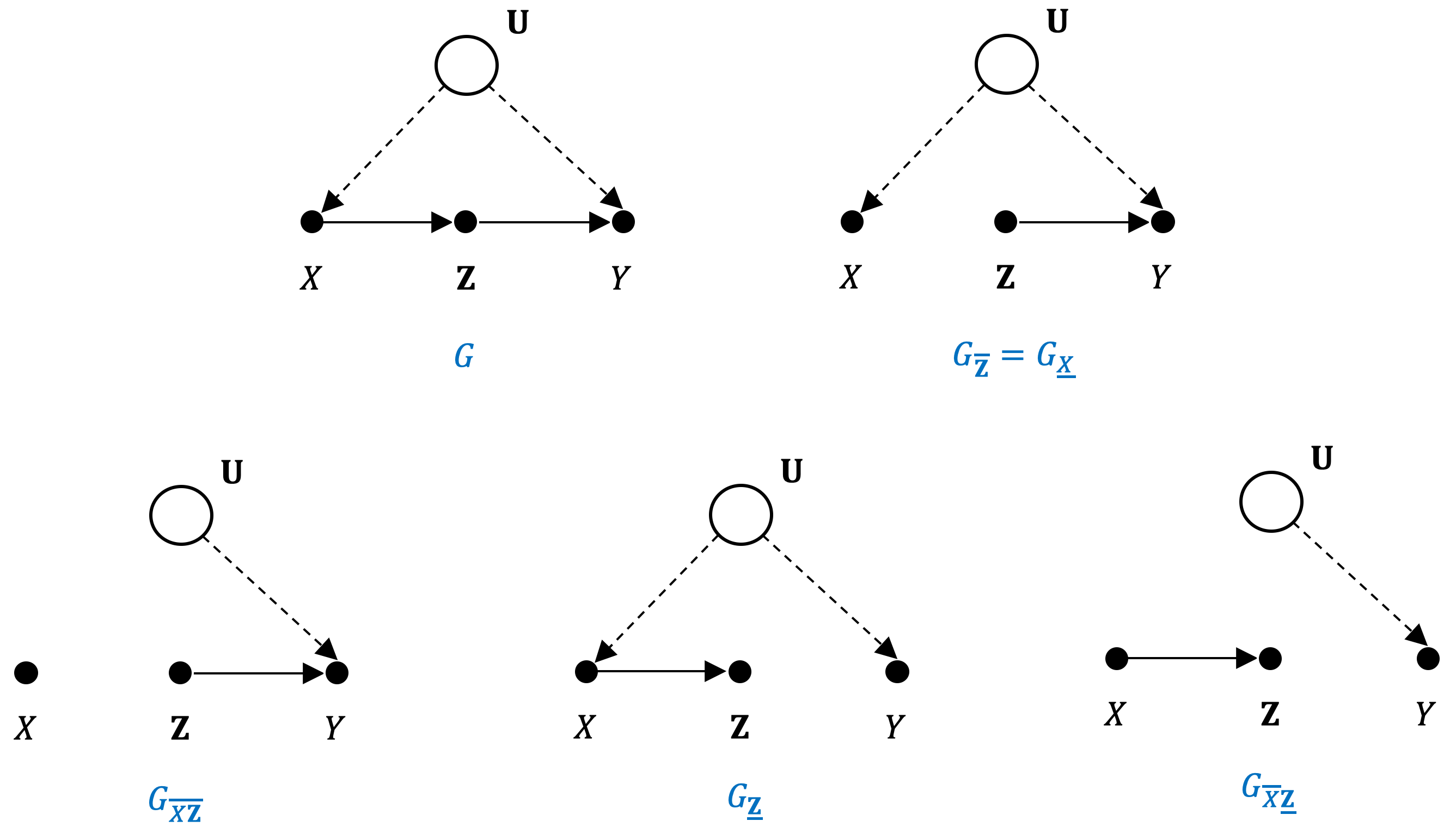}
\caption{\small{The original graph $G$ considered in the front-door rule and the subgraphs used for its derivation.}}
\label{front_door_derivation_fig}
\end{figure}

As an example of an application of the do Calculus let's see how we can use them to derive the front door rule (\ref{front_door_eq}), referring to figure \ref{front_door_derivation_fig} for the various subgraphs that we shall need to examine.
\\
Lets go through this step by step following \cite{EMP}.
\begin{itemize}

\item First, using the sum rule and definition (\ref{mixed_cond_eq}), the desired quantity can be written as:
\begin{equation}
	p(\mathbf{y}|\:\Do(\mathbf{x})) = 
	\sum_{\mathbf{z}} p(\mathbf{y},\mathbf{z}|\:\Do(\mathbf{x})) = 
	\sum_{\mathbf{z}} p(\mathbf{y}|\:\mathbf{z},\Do(\mathbf{x}))\:p(\mathbf{z}|\:\Do(\mathbf{x})).
	\label{step_1_eq}
\end{equation}

\item Now we can apply rule 3 to remove the do operator on $\mathbf{x}$ in the second factor of (\ref{step_1_eq}) to write 
\begin{equation}
	p(\mathbf{z}|\:\Do(\mathbf{x}))=p(\mathbf{z}|\mathbf{x})
	\label{step_2_eq}
\end{equation}

because $(\mathbf{Z} \independent \mathbf{X})_{G_{\underline{\mathbf{X}}}}$ as $Y$ is an unconditionned collider in $G_{\underline{\mathbf{X}}}$.

\item Lets now focus on the factor $p(\mathbf{y}|\:\mathbf{z},\Do(\mathbf{x}))$ in (\ref{step_1_eq}). Using rule 3 again on $G_{\overline{\mathbf{X}}\underline{\mathbf{Z}}}$ which satisfies $(\mathbf{Y}\independent\mathbf{Z}|\mathbf{X})_{G_{\overline{\mathbf{X}}\underline{\mathbf{Z}}}}$, because there is no path at all between $\mathbf{Y}$ and $\mathbf{Z}$ in $G_{\overline{\mathbf{X}}\underline{\mathbf{Z}}}$, we can replace $\mathbf{z}$ by $\Do(\mathbf{z})$
\begin{equation}
	p(\mathbf{y}|\:\mathbf{z},\Do(\mathbf{x})) =
	p(\mathbf{y}|\:\Do(\mathbf{z}),\Do(\mathbf{x})).
	\label{step_3_eq}
\end{equation}

\item Rule 2 now allows deleting the $\Do(\mathbf{x})$ action from the right member of (\ref{step_3_eq}) because $(\mathbf{Y}\independent\mathbf{X}|\:\mathbf{Z})_{G_{\overline{\mathbf{X}}\overline{\mathbf{Z}}}}$, therefore
\begin{equation}
	p(\mathbf{y}|\:\Do(\mathbf{z}),\Do(\mathbf{x}))=p(\mathbf{y}|\:\Do(\mathbf{z})).
	\label{step_4_eq}
\end{equation}
\item Lets thus compute $p(\mathbf{y}|\:\Do(\mathbf{z}))$. Apply definition (\ref{mixed_cond_eq}) on mixed conditioning and marginalizing over $\mathbf{x}$
\begin{equation}
	p(\mathbf{y}|\:\Do(\mathbf{z})) =
	\sum_{\mathbf{x}} p(\mathbf{y},\mathbf{x}|\:\Do(\mathbf{z})) =
	\sum_{\mathbf{x}} p(\mathbf{y}|\mathbf{x},\Do(\mathbf{z}))\:
	p(\mathbf{x}|\Do(\mathbf{z})).
	\label{step_5_eq}
\end{equation}
We can apply rule 2 which implies $p(\mathbf{x}|\Do(\mathbf{z}))=p(\mathbf{x})$ because $(\mathbf{X}\independent\mathbf{Z})_{G_{\overline{\mathbf{Z}}}}$, the path between $\mathbf{X}$  and $\mathbf{Z}$ being blocked by $\mathbf{Y}$ which is an unconditionned collider. We can also apply rule 3 which implies $p(\mathbf{y}|\mathbf{x},\Do(\mathbf{z}))=p(\mathbf{y}|\mathbf{x},\mathbf{z})$ because $(\mathbf{Y}\independent\mathbf{Z}|\mathbf{X})_{G_{\underline{\mathbf{Z}}}}$ because $\mathbf{X}$ is an observed fork which blocks the path between $\mathbf{Y}$  and $\mathbf{Z}$. Altogether (\ref{step_5_eq}) and these last two results imply
\begin{equation}
	p(\mathbf{y}|\:\Do(\mathbf{z})) = \sum_{\mathbf{x}'} 
	p(\mathbf{y}|\mathbf{x}',\mathbf{z}) \:
	p(\mathbf{x}').
	\label{step_6_eq}
\end{equation}
Combining (\ref{step_3_eq}), (\ref{step_4_eq}) and (\ref{step_6_eq}), we get
\begin{equation}
	p(\mathbf{y}|\:\mathbf{z},\Do(\mathbf{x})) = 
	\sum_{\mathbf{x}'} 
	p(\mathbf{y}|\mathbf{x}',\mathbf{z}) \:
	p(\mathbf{x}').
	\label{step_7_eq}
\end{equation}
\item Eventually, combining (\ref{step_1_eq}), (\ref{step_2_eq}) and (\ref{step_7_eq}), we get the desired result
\begin{equation}
	p(\mathbf{y}|\:\Do(\mathbf{x})) = \sum_{\mathbf{z}}
	p(\mathbf{z}|\mathbf{x})\:
	\sum_{\mathbf{x}'} 
	p(\mathbf{y}|\mathbf{x}',\mathbf{z}) \:
	p(\mathbf{x}').
\end{equation}
which is the same as the front door rule (\ref{front_door_eq}).
\end{itemize}
As this example shows, the do Calculus allows for elegant derivations of identifiability cases but by no means is it obvious! On the contrary, it requires a fair amount of intuition and practice.

\subsubsection*{Strengths and Limitations of the Do Calculus}

\begin{figure}[h]
\centering
\includegraphics[scale=0.25]{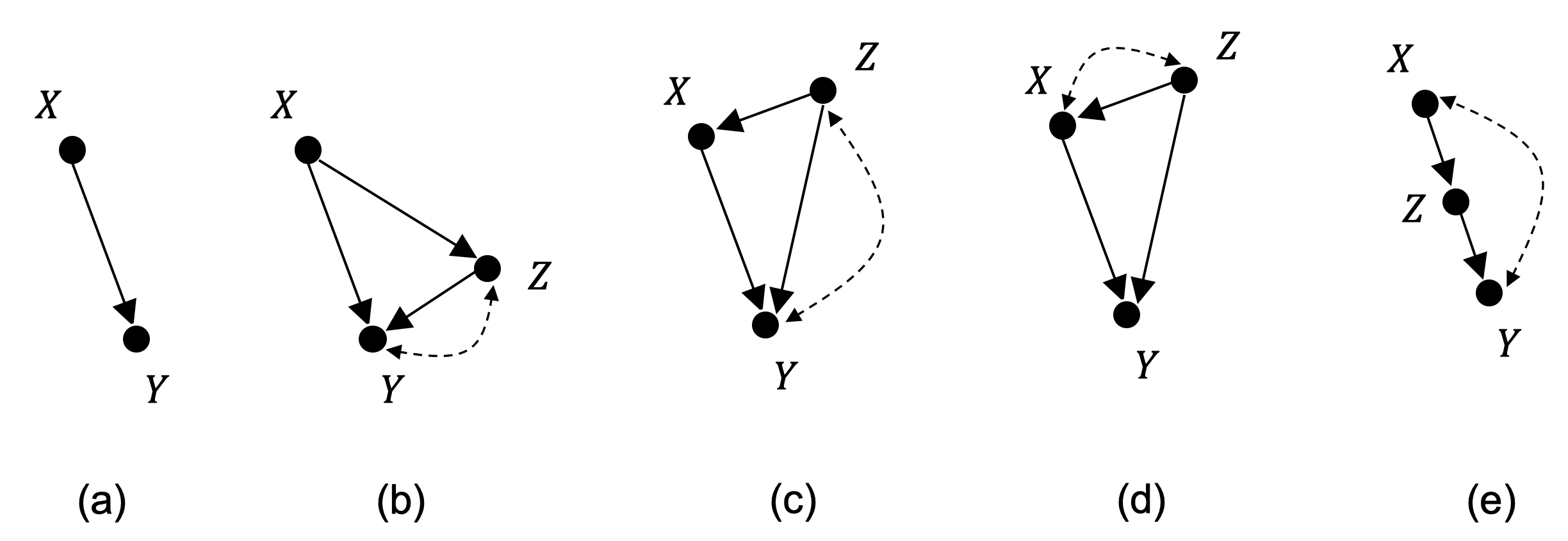}
\caption{\small{Examples of simple causal graphs where the effect of $X$ on $Y$ is identifiable. All graphs are maximal.}}
\label{identifiable_fig}
\end{figure}

The rules of the do Calculus have been shown to be \textbf{complete} \cite{COMPLETENESS} in the sense that, if an intervention is identifiable at all, then there exists indeed a sequence of these 3 rules that will produce an expression free of any do operator. Unfortunately, there is \textbf{no general rule} for deciding whether an expression is identifiable or not \cite{CAU}. In fact one of the main difficulty in applying the do calculus is that there is no general guiding principle that tells us which rule to apply at each step. Luckily, courageous people have created lists of identifiable and non identifiable causal diagrams that we can reuse. Figure \ref{identifiable_fig} shows examples which are identifiable. 
\\
One important remark is that we can generate many more identifiable diagrams from these as soon as we realize the following two facts:
\begin{figure}[h]
\centering
\includegraphics[scale=0.25]{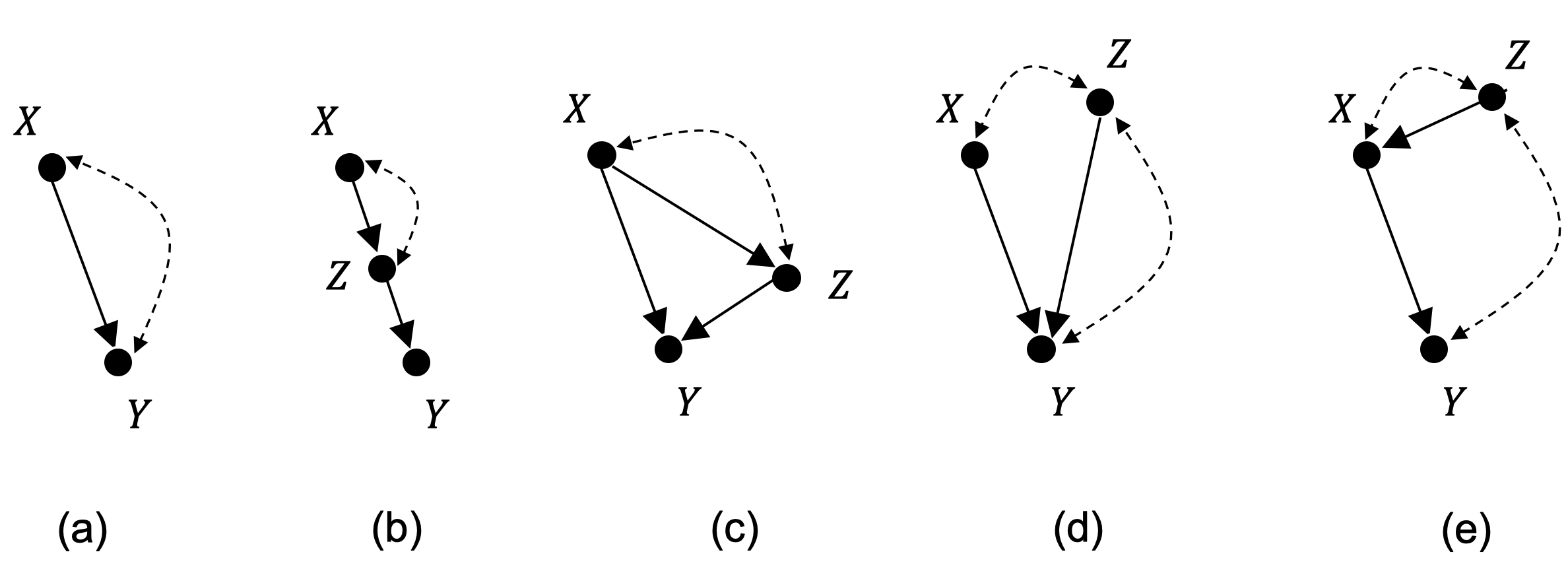}
\caption{\small{Examples of simple causal graphs where the effect of $X$ on $Y$ is not identifiable.}}
\label{unidentifiable_fig}
\end{figure}
\begin{itemize}

\item \textbf{Adding an edge} to any causal diagram $G$ can only decrease the chances that the new graph will be identifiable. This is a direct consequence of the fact that all rules we have been discussing, whether graphical or algebraic, ultimately amount to testing whether some paths within subsets of $G$ are blocked. Adding an edge obviously increases the number of paths to test and thus decreases the chances that they will all be blocked.

\item \textbf{Adding an observed variable} on an edge of a graph has the opposite effect. It can only increase the chances that the new graph will be identifiable because it adds an observed chain which is an additional blocking point that can only enhance the chances of  $d$-separation between groups or variables.

\end{itemize}
All diagrams shown in figure \ref{identifiable_fig} are \textbf{maximal} in the sense that adding a single edge to any of them will destroy their identifiability. They are thus the most useful graphs to list. Figure \ref{unidentifiable_fig} shows examples which are not identifiable.
\section{Causal Analysis of Simple Business Examples}
\label{examples_sec}
Examples of causal analysis are often drawn from fields like agronomy, social sciences or economics which rely on complex causal models. We prefer to illustrate how the need for causal analysis comes naturally in simple business examples.

\subsection{Assessing the Effectiveness of a Loyalty Campaign}
The level of satisfaction of customers with a service can often be judged from their observable behavior such as the number of visits to a website or the time spent for using that service. This information can help the service provider to identify early signs that some customers intend to leave the service. Marketing can then decide to launch a loyalty campaign that aims at retaining the customers whose faith wavers. These campaigns of course have a cost and its therefore important to be able to assess their effectiveness. Does a loyalty campaign of a certain type really have an impact on the attrition rate? 
\\

\begin{figure}[h]
\centering
\includegraphics[scale=0.25]{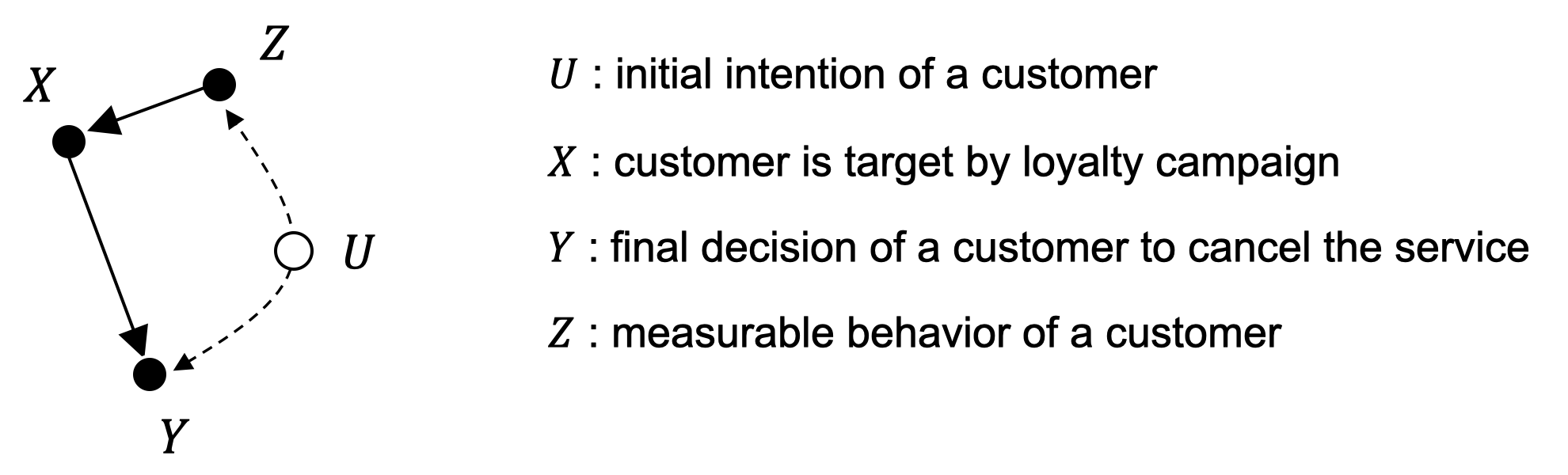}
\caption{\small{The causal model of the impact of a loyalty campaign on the churn rate.}}
\label{churn_fig}
\end{figure}

To answer this question let's formalize the problem. First, let $U$ be a binary variable that corresponds to the intention of the customer to leave the service or not, before being targeted by any loyalty campaign. This, of course, will be an unmeasured variable. This initial intention $U$ certainly has an impact on the behavior of the customer that can be measured, let's call this $Z$. This behavior $Z$ will determine in turn the decision by the service provider whether this customer should be targeted by a loyalty campaign. Let's call this decision $X$. Eventually, the final decision of the customer to cancel the service is again a binary variable, call it $Y$. This final decision of the customer is certainly dependent both on his initial intention $U$ and whether he has been targeted or not by the campaign ($X$). So, formally, our question translates as : can we identify the causal impact of having been a target $X$ on the final decision $Y$ when we measure $Z$ ? Figure \ref{churn_fig} shows the corresponding causal diagram. 
\\
This diagram is nothing but the diagram (c) from the list of identifiable diagrams in figure \ref{identifiable_fig} with the direct link from $Z$ to $Y$ removed. Therefore it is identifiable. On the other hand $Z$ satisfies the two condition of the back-door criterion in theorem \ref{backdoor_theorem}. It is not a descendant of $X$ and it blocks the back-door path between $X$ and $Y$ because it is an observed chain. Thus (\ref{backdoor_reduction}) gives the explicit reduction of the causal impact of $X$ on $Y$ into observable quantities.

\subsection{Claim Provisioning by an Insurance Company}
When a claim is reported by an insured, an insurer usually sets aside a certain amount. This is especially true when personal injury is reported. The final compensation will obviously depend on the extent of the damage. But, as insurers have become aware, this compensation will also depend on the initial provisioning. The aim of causal analysis here is to determine the impact of this provisioning on the final compensation granted to the insured.
\\

Let $U$ be the complete claim characterization. Its value is unknown at the time of reporting but it will determine both the declaration $Z$ by the insured and the final compensation $Y$ granted after investigations. The initial evaluation $Z$ is what directly determines the expert's provisioning $X$ which in turns impacts the final compensation $Y$. The causal graph is thus exactly the same as in figure \ref{churn_fig} and, as a consequence, the same back-door reduction formula (\ref{backdoor_reduction}) applies here as well.

\subsection{Impact of Sales Force Training on Turnover}
\begin{figure}[h]
\centering
\includegraphics[scale=0.25]{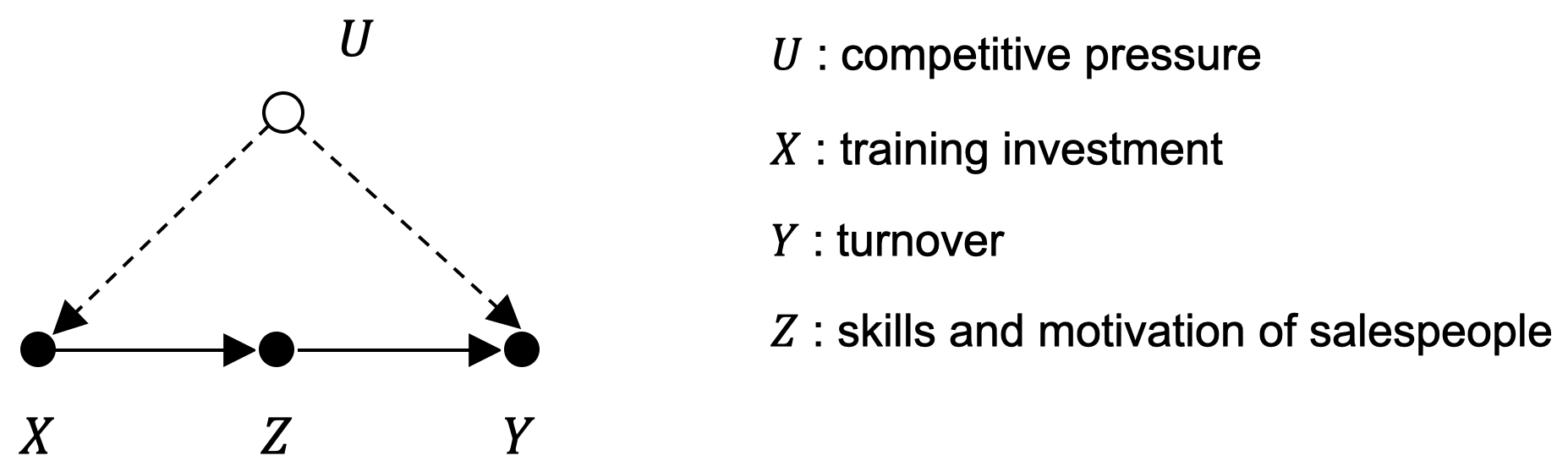}
\caption{\small{Causal graph for the impact of a training investment on the turnover leveraging an evaluation of the skills and motivation of salespeople.}}
\label{sales_force_fig}
\end{figure}

To improve the efficiency of its sales teams, a company can invest in their training. This has an easily measurable cost $X$ which it is hoped will help to increase the turnover $Y$ by increasing sales. As such, the causal relationship would be direct and thus trivially identifiable. However there is at least one unmeasured confounding variable which spoils this identifiability, namely the competitive pressure $U$. Indeed a strong competitive pressure $U$ will encourage investment $X$ in a training effort but, simultaneously, it will also impact the turnover $Y$. Fortunately though, we can restore identifiability if we can find a measurable ``deconfounding'' variable $Z$ within the causal chain from $X$ to $Y$. The skills and motivation of salespeople induced by these training and measured by surveys or polls could play this role. The causal graph is depicted in figure \ref{sales_force_fig}, it is the same as in figure \ref{front_door_fig} which was the basic graph for which the front-door rule applies. We can thus use the front-door reduction formula (\ref{front_door_eq}).

\subsection{Impact of Pricing on Sales}
\begin{figure}[h]
\centering
\includegraphics[scale=0.28]{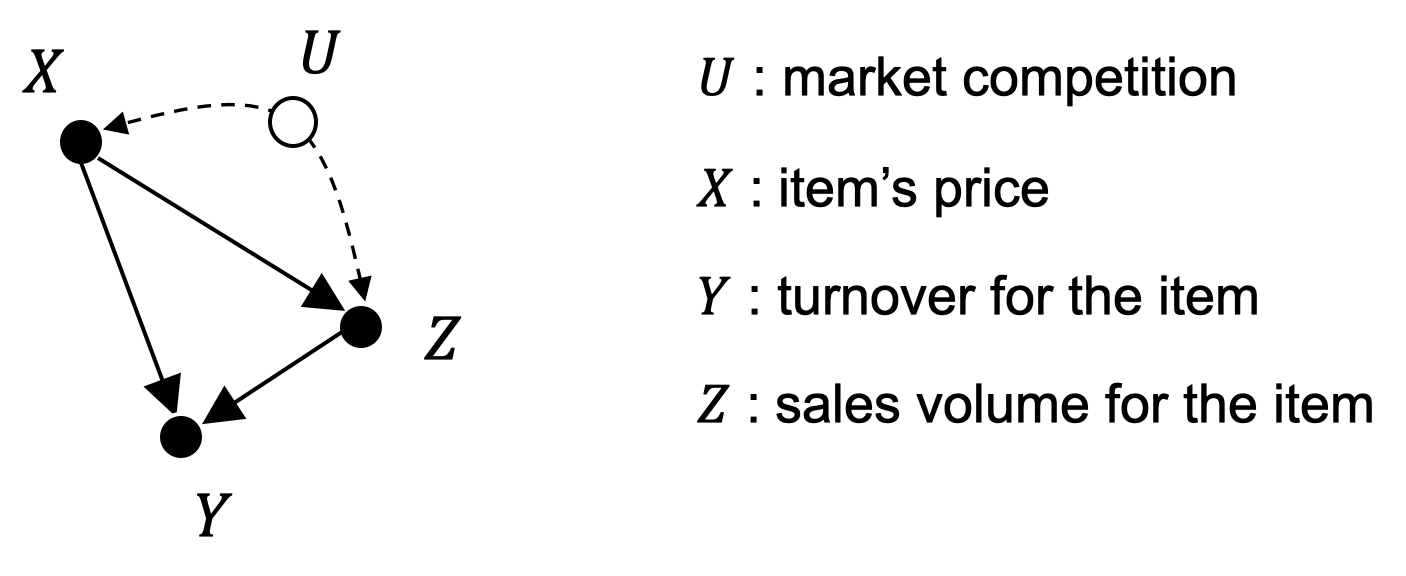}
\caption{\small{The impact of pricing on turnover is unidentifiable because of the competition pressure.}}
\label{price_on_sales_fig}
\end{figure}

The turnover $Y$ for a company on a product is a function of the price $X$ of the item and its sales volume $Z$. For instance $Y$ could be proportional to $X \cdot Z$. Now the price $X$ impacts both the turnover $Y$ but also the sales volume $Z$, because lower prices generally increasing the volume. As such, the corresponding causal graph would be identifiable because it corresponds to graph (b) from in the list of identifiable graphs from figure \ref{identifiable_fig} with the confounding link between $Y$ and $Z$ removed. Unfortunately the competition pressure $U$ which is an unknown variables is a confounder which impacts both the price $X$ and the sales volume $Z$ rendering the graph shown in figure \ref{price_on_sales_fig} unidentifiable as it is the same as the graph (c) in the list of unidentifiable graphs in figure \ref{unidentifiable_fig}.
\section{Stuff Worth Knowing for Data Scientists?}
Once one fully grasps the importance of the paradigm shift implied by the Causal Revolution, a question will occur to many newcomers. Why hasn't anyone ever told me about this while I as learning statistics, probability or machine learning? How come that as a data scientist I've never heard of the do Calculus? Is it really useful in practice? In what area? There are probably several answers to this. We list some of them below.
\begin{itemize}

\item The first reason we see is of cultural nature. This  has been well analyzed in \cite{WHY_REV} for instance. In reality the word ``causality'' has long remained almost taboo in the statistical community which remained faithful to the classical associationist point of view of the founders of the discipline in the last century like Pearson and Fisher. They considered that there can be no evidence about causation within data (except for RCT). It took all the perseverance and scientific stature of  Judea Pearl to reverse people's minds on the relationship between causality and statistics. Apparently, this paradigm shift has yet to infuse the newer data science community at large.

\item The graphical and algebraic machinery we presented in section \ref{CR_in_Nutshell_section} obviously only makes sense when the causal graph $G$ is non trivial. This explains why fields like social sciences, economics, medicine or agronomy, which have all developed a rich set of causal models, have benefited most from this renewed perspective on causality. For an elementary causal relationship like $G=X \rightarrow Y$ there is however no difference between the passive observation of the cause $X$ and an intervention on it. Many use cases worked out by data scientist fall into this category, perhaps in some cases just by chance. But without such luck, neglecting the difference between observation and intervention could lead to predictions errors. We presented a number of examples of such cases in section \ref{examples_sec}.

\item While the graphical rules are rather accessible, the judicious application of the do Calculus to new causal diagrams undoubtedly requires skill and intuition. As was suggested in \cite{WHY_REV} those who cannot afford or don't want to master this kind of algebraic machinery could certainly benefit from some sort of cookbook.

\item One last reason, a more subjective one, has to do with Pearl's writing style \cite{CAU} which, at times lacks perhaps a bit of conciseness. This can certainly discourage many readers, even those those who have all prerequisites in statistics and probability theory. The sheer number of historical anecdotes, examples and comments that Pearl gratifies his readers with sometimes runs counter to a clear presentation of the key mathematical facts. The Causality Revolution probably still lacks a concise textbook like presentation of the subject.
 
\end{itemize}

The founding fathers of statistics were certainly right when they insisted that, in general, we cannot infer causal relations from mere data. But in the enthusiasm of their nascent discipline, many of their disciples and epigones have somehow become overzealous. They neglected that in most cases we have more information available than just data! We have some prior knowledge of how the world works! Perhaps not a full-fledged causal model, but at least a partial causal graph. What Pearl tells us is that in some favorable conditions (that we discussed in section \ref{CR_in_Nutshell_section}), we can merge those two pieces of knowledge in order to predict the effect of an intervention on a system. If it turns out that these predictions do not match observations, we just have to work harder and think of some better model. But, in any case, we have to give up the hope of automating the discovery of how the world works from observation only. Nature won't reveal its charms in front of some dumb algorithm. She is a demanding mistress and expects us to engage in a thoughtful conversation with her, as curious and creative minds. So, let's rejoice, because this is why science is so much fun!

\end{document}